\documentclass{article}

    \usepackage[preprint]{neurips_2022}

\usepackage[utf8]{inputenc} 
\usepackage[T1]{fontenc}    
\usepackage{hyperref}       
\usepackage{url}            
\usepackage{booktabs}       
\usepackage{amsfonts}       
\usepackage{nicefrac}       
\usepackage{microtype}      
\usepackage{xcolor}         
\usepackage{graphicx}       
\usepackage{optidef}
\usepackage{caption}
\usepackage{subcaption}
\usepackage{amsthm}
\usepackage{amssymb}

\def\thmref#1{theorem~\ref{#1}}
\def\Thmref#1{Theorem~\ref{#1}}
\def\propref#1{proposition~\ref{#1}}
\def\Propref#1{Proposition~\ref{#1}}

\def\eqnref#1{equation~\eqref{#1}}
\def\Eqnref#1{Equation~\eqref{#1}}
\def\optref#1{program~\eqref{#1}}

\def\figref#1{figure~\ref{#1}}
\def\Figref#1{Figure~\ref{#1}}
\def\secref#1{section~\ref{#1}}

\def\lemref#1{lemma~\ref{#1}}
\def\Lemref#1{Lemma~\ref{#1}}

\newcommand{\seq}[3]{
    \ensuremath{\{#1\}_{#2}^{#3}}
}
\newcommand{\XX}[0]{\mathcal{X}}
\newcommand{\HH}[0]{\mathcal{H}}
\newcommand{\SSS}[0]{\mathcal{S}}
\newcommand{\CC}[0]{\mathcal{C}}

\newcommand{\NN}[0]{\mathcal{N}}

\newcommand{\AAA}[0]{\mathcal{A}}
\newcommand{\ann}[2]{
    \ensuremath{
        \NN\big(#1,#2\big)
    }
}

\theoremstyle{plain}
\newtheorem{theorem}{Theorem}[section]
\newtheorem{proposition}[theorem]{Proposition}
\newtheorem{lemma}[theorem]{Lemma}

\theoremstyle{definition}
\newtheorem{definition}[theorem]{Definition}

\theoremstyle{remark}

\title{An Analytic Framework for Robust Training of \\ Artificial Neural Networks}

\author{%
  Ramin Barati \\
  Department of Computer Engineering\\
  Amirkabir University of Technology\\
  Tehran, Iran \\
  \texttt{ramin.barati@aut.ac.ir} \\
  \And
  Reza Safabakhsh \\
  Department of Computer Engineering\\
  Amirkabir University of Technology\\
  Tehran, Iran \\
  \texttt{safa@aut.ac.ir} \\
  \AND
  Mohammad Rahmati \\
  Department of Computer Engineering\\
  Amirkabir University of Technology\\
  Tehran, Iran\\
  \texttt{rahmati@aut.ac.ir} \\
}

\begin{document}

\maketitle

\begin{abstract}
The reliability of a learning model is key to the successful deployment of machine learning in various industries. Creating a robust model, particularly one unaffected by adversarial attacks, requires a comprehensive understanding of the adversarial examples phenomenon. However, it is difficult to describe the phenomenon due to the complicated nature of the problems in machine learning. Consequently, many studies investigate the phenomenon by proposing a simplified model of how adversarial examples occur and validate it by predicting some aspect of the phenomenon. While these studies cover many different characteristics of the adversarial examples, they have not reached a holistic approach to the geometric and analytic modeling of the phenomenon. This paper propose a formal framework to study the phenomenon in learning theory and make use of complex analysis and holomorphicity to offer a robust learning rule for artificial neural networks. With the help of complex analysis, we can effortlessly move between geometric and analytic perspectives of the phenomenon and offer further insights on the phenomenon by revealing its connection with harmonic functions. Using our model, we can explain some of the most intriguing characteristics of adversarial examples, including transferability of adversarial examples, and pave the way for novel approaches to mitigate the effects of the phenomenon.
\end{abstract}

\section{Introduction}
\label{sec:intro}
The state-of-the-art neural models are shown to suffer from the phenomenon of adversarial examples, where a machine learning model is fooled to return an undesirable output on particular inputs that an adversary carefully crafts. The phenomenon is peculiar because it seems to affect neural networks in every application globally and that the adversarial examples are invariant to changes in network architecture, transferring from one network to another. From the perspective of learning theory, the existence of these samples is paradoxical since the nonrobust networks show acceptable, even super-human, performance on the natural samples. In the literature, many attempts at resolving this paradox have been made, each revealing a different facet of the phenomenon.

\cite{szegedy2013intriguing} explained the adversarial examples as small pockets in the domain of the hypothesis, where the hypothesis fails to be correct due to its highly nonlinear nature. In contrast, \cite{goodfellow2014explaining} proposed that the phenomenon is a side-effect of a linear hypothesis in high dimensions. Conversely, \cite{ilyas2019adversarial} blamed useful nonrobust features that are effective in dealing with natural samples; but are a hindrance when the model is tested on adversarial examples. \cite{9412367} proposes that all of these different and opposing perspectives unite under the banner of pointwise convergence of the hypothesis to the optimal hypothesis. On a separate thread, \cite{tanay2016boundary,shamir2021dimpled} propose geometrical descriptions of the phenomenon in which adversarial examples are attributed to gemetrical interactions between the manifold of natural samples and the decision boundary. 

In this paper, we take the first steps for the analysis the phenomenon through the lens of complex analysis. The use of complex variables and functions enables us to consider the algebraic and the geometric properties of the phenomenon simultaneously, leading to a rigorous framework to study the phenomenon and a possible remedy for its effects. For this reason, we introduce a novel approach in describing the phenomenon using complex-valued hypotheses. We then introduce the space of holomorphic hypotheses and show that, in the limit of infinite samples, all holomorphic hypotheses are forced to converge to the same holomorphic hypotheses, explaining the transfer of adversarial examples between analytic hypotheses. Finally, we will generalize the results to real-valued hypothesis through the methods of calculus of variation and differential equations, and derive a robust learning rule for artificial neural networks.

\section{Preliminaries}
\label{sec:pre}
In this section, we provide a summary of the relevant concepts in learning theory, and formally describe the objects of our study. We refer the interested reader to \cite{shalev2014understanding} for a through discussion of learning theory. A goal of learning theory is to provide a way to decide on the learnability of a hypothesis class $\HH$ and quantize the hardness of a learning task for a learning rule $\AAA$ with respect to $\HH$. The complexity of a learning task is decided by its sample complexity function. The sample complexity is an integer-valued function that determines a lower bound on the size of the training set relative to a desired error magnitude $\epsilon$ and an error probability $\delta$. The training set is sampled from a domain set $\XX$ and is labeled by a labeling function $t:\XX\rightarrow\{+1,-1\}$.

A restricted and strong form of learning is the probably approximately correct (PAC) learning. PAC learnability of $\HH$ means that a learning rule $\AAA$ exists that its sample complexity with respect to $\HH$ is only dependent on $\epsilon$ and $\delta$. In other words, not matter the task, the training set size is only proportional to the desired generalization error. PAC learnability and uniform convergence are related through the fundamental theorem of statistical machine learning.

A more relaxed notion of learnability is nonuniform learnability. The sample complexity in nonuniform learning is dependent on the optimal hypothesis. In other words, some hypotheses need a larger training set for $\AAA$ to successfully pass the learning task requirements.

It is possible to further weaken and relax nonuniform learnability and
allow $\AAA$ to be consistent with respect to $\HH$. Consistency is characterized by the dependency of the sample complexity on the sample distribution. Consequently, the learner needs an even larger training set to deal with complications in sampling from $\XX$ on top of the complexity of a complicated learning goal. Consistency and pointwise convergence are related through the concept of convergence in probability.

The first account for the existence of optimal training sets and their relation with adversarial examples phenomenon is due to \cite{ilyas2019adversarial}. It was later shown by \cite{9412367} that the output of a consistent learning rule could be weak against adversarial attacks in simple learning tasks. Consequently, we argue that if $\AAA$ realizes nonuniform learnability with respect to $\HH$, then $\AAA$ is a robust learner with respect to $\HH$.

\begin{definition}[uniform convergence]
A sequence of functions $\seq{f_n:\XX\rightarrow\mathbb{R}}{n=1}{\infty}$ is uniformly convergent to $f:\XX\rightarrow\mathbb{R}$ if for every $\epsilon>0$, there exists a natural number $N$ such that for all $n\geq N$ and $x\in\XX$ we have that
\begin{equation}
    |f_n(x)-f(x)|\leq\epsilon.
\end{equation}
\end{definition}

Uniform convergence of a sequence of functions is a strong condition on the sequence in general. For example, the sequence $\seq{x^n}{n=1}{\infty}$ does not converge uniformly to $f(x)=0$ on the open interval $\XX=(0,1)$. Thus, other modes of convergence have been introduced for analysis of functions.

\begin{definition}[compact convergence]
A sequence of functions $\seq{f_n:\XX\rightarrow\mathbb{R}}{n=1}{\infty}$ is compactly convergent to $f:\XX\rightarrow\mathbb{R}$ if it is uniformly convergent on every compact subset $K\subseteq\XX$. 
\end{definition}

In this paper, we relax the notion of uniform convergence to compact convergence so that our analysis is not hindered by some complications in the choice of $\XX$.

\begin{definition}[robust learner]
\label{dfn:robust learner}
A learning rule $\AAA$ is robust with respect to $\HH(\XX)$ if for all $\epsilon\geq 0$ a natural number $N$ exists in which for all training sets $S,S'\subseteq\XX$ larger than $N$ we have that
\begin{equation}
    \sup_\XX|\AAA(S')(x)-\AAA(S)(x)|\leq\epsilon.
\end{equation}
\end{definition}

This definition of a robust learner could be interpreted as the characteristic that when the learner observes enough natural samples, adding more samples would not be consequential to the learners output, independent from the generative process of the samples.

\begin{theorem}
\label{thm:robustness}
If a learning rule $\AAA$ is a nonuniform learner with respect to $\HH$, then it is robust.
\end{theorem}

Given that the main focus of this paper is on artificial neural networks (ANNs), we need an abstraction for ANNs that is susceptible to analysis under this framework. The first difficulty is that ANNs are trained using gradient decent, which would produce an infinite subsequence for every training set in our sequence. The second difficulty is that activation functions in ANNs can become quite complicated; e.g. they could be constructed with the help of other ANNs. To simplify these aspects of analysis of ANNs, we introduce an abstraction of ANNs.

\begin{definition}[generalized ANN]
The hypothesis class of generalized ANNs $\ann{\HH(\Omega)}{\sigma(\XX)}$ consists of functions from a domain set $\XX$ to either $\mathbb{C}$ or $\mathbb{R}$ which has a representation of the form
\begin{equation}
\label{eqn:form}
    f(x)=\int_\Omega h(\omega)s(x;\omega)\,dV(\omega) \quad h\in \HH(\Omega),s\in\sigma(\XX)
\end{equation}
in which $dV(\omega)$ is the volume differential of $\Omega$, $\HH(\Omega) \subseteq L^2(\Omega)$ is a hypothesis class on the domain set $\Omega$, and $\sigma(\XX) \subset L^2(\XX)$ is a family of activation functions on $\XX$ parameterized by $\omega\in\Omega$.
\end{definition}

\begin{definition}[universal approximation property]
$\sigma(\XX)$ has the universal approximation property if for all $\alpha\in\mathbb{N}$ an $\omega\in\Omega$ could be found where
\begin{equation}
    \int_\XX\varphi_\alpha(x)s(x;\omega)\,dV(\XX)\neq0,
\end{equation}
in which $\seq{\varphi_\alpha}{\alpha=0}{\infty}$ is the complete orthonormal basis for $L^2(\XX)$.
\end{definition}

\begin{lemma}
\label{lem:subset}
$\ann{\HH(\Omega)}{\sigma(\XX)}\subseteq\HH(\XX)$ with equality being true when $\sigma(\XX)$ has the universal approximation property.
\end{lemma}

The largest hypothesis class that we consider here is $L^2(\XX)$. The main characteristic of functions in $L^2(\XX)$ is that they are square-integrable. Formally, for a function $f\in L^2(\XX)$,
\begin{equation}
    \|f\|_{L^2(\XX)}=\Big(\int_\XX |f(x)|^2\,dV(x)\Big)^\frac{1}{2}<\infty.
\end{equation}

\begin{theorem}
\label{thm:non learnability}
$L^2(\XX)$ is nonuniform learnable.
\end{theorem}

\begin{definition}[SVC learning rule]
Consider a training set $S=\seq{(x_n\in\XX,t(x_n))}{n=1}{N}$ and a hypothesis $f\in\ann{L^2(\Omega)}{\sigma(\XX)}$. The support vector classifier (SVC) learner solves the following program,
\begin{argmini}
{v}{\frac{1}{2}\|h\|^2_{L^2(\Omega)}}
{\label{opt:lmm}}{}
\addConstraint{t_nf(x_n)}{\geq 1}
\end{argmini}
\end{definition}

\begin{theorem}
\label{thm:svc consistency}
The SVC learning rule is not nonuniform with respect to $\ann{L^2(\Omega)}{\sigma(\XX)}$, and the optimal solution $h^*$ is,
\begin{equation}
    h^*(\omega)=\sum_{n=1}^N\lambda_nt_n\sigma(x_n;\omega),
\end{equation}
in which $\seq{\lambda_n}{n=1}{N}$ are the Lagrange multipliers of \optref{opt:lmm}.
\end{theorem}

Considering \thmref{thm:svc consistency}, we expect that the output of the SVC learner would be weak against adversarial attacks in general. 

\begin{proposition}
\label{prop:l2 transfer}
The SVC learner of a hypothesis class $\ann{\HH(\Omega)}{\sigma(\XX)}$ would return the labeling function $t$ as the output in the infinite sample limit unless it is agnostic to $t$.
\end{proposition}

Based on \propref{prop:l2 transfer}, one may suggest that SVC learners with respect to $L^2(\XX)$ are weak against black box attacks as well. We see that if we want to take a full account of the adversarial examples phenomenon, we need to formalize the conditions for transferability of adversarial examples. \cite{236234} have proposed three metrics for measuring transferability between a target $f\in\HH(\XX)$ and a surrogate $\hat{f}\in\hat{\HH}(\XX)$. The first metric is the size of the gradient of the loss function with respect to input $\|\nabla_x\ell\|$ for the target hypothesis. $\|\nabla_x\ell\|$ basically measures the robustness of the target model. The second metric is the cosine distance between the gradient of the loss of the target $\nabla_x\ell$ and of the surrogate $\nabla_x\hat{\ell}$. In other words, how aligned the gradients are. The third metric is the variance of the loss landscape of the surrogate hypothesis class
\begin{equation}
\mathbb{E}_\XX\big[\ell\big(t(x),x,\hat{f}\big)^2\big]-\mathbb{E}_\XX\big[\ell\big(t(x),x,\hat{f}\big)\big]^2,    
\end{equation}
which compensates for the fact that the surrogate is not trained on the same training set as the target. Inspired by the definition of a normal family of functions, we introduce normal hypothesis classes.

\begin{definition}[normal hypothesis class]
\label{def:normal}
$\HH(\XX)$ is a normal hypothesis class if for every universally consistent learning rule $\AAA$ and all $\epsilon,\delta \geq 0$ a number $N\in\mathbb{N}$ exists where for all sets of training points $S\subseteq\XX$ with size larger than $N$ we have with probability $1-\delta$ that
\begin{equation}
    |\AAA(S)(x)-\AAA(\XX)(x)|\leq\epsilon.
\end{equation}
\end{definition}

Based on the aforementioned metrics, we can see that if both $\HH(\XX)$ and the class of its derivatives $\partial\HH(\XX)$ are normal, and if $\hat{f}\in\HH(\XX)$, then an attacker would know that the adversarial examples of $\hat{f}$ would transfer to a nonrobust $f$ up to a probability.

\begin{definition}[transfer of adversarial examples]
The adversarial examples of $\hat{\AAA}(\hat{S}_n)=\hat{f}_n\in\hat{\HH}(\XX)$ would generalize up to a probability to $\AAA(S_m)=f_m\in\HH(\XX)$ if for all $\epsilon,\epsilon',\delta \geq 0$ some natural number $N$ exists in which for all $n,m \geq N$ it is true that with probability $1-\delta$
\begin{align}
|\hat{f}_n-f_m|&\leq\epsilon, \\
\|\nabla\hat{f}_n-\nabla f_m\|&\leq\epsilon'
\end{align}
\end{definition}
The idea behind this definition of transfer is that not only $\hat{f}$ and $f$ vary in a small neighborhood of the infinite sample limit, they also vary in a similar manner in that neighborhood.
\begin{theorem}
\label{thm:ann transfer}
Consider a nonrobust target $f\in\ann{\HH(\Omega)}{\sigma_1(X)}$ and a surrogate $\hat{f}\in\ann{\HH(\Omega)}{\sigma_2(X)}$. If $\HH(\Omega)$ and $\partial\HH(\Omega)$ are normal and $\sigma_1(\XX)$ and $\sigma_2(\XX)$ have the universal approximation property, then the adversarial examples of $\hat{f}$ will transfer to $f$.
\end{theorem}

\begin{theorem}
\label{thm:not normal}
$L^2(\XX)$ is not a normal hypothesis class.
\end{theorem}

Even though \thmref{thm:non learnability} shows that it is possible to find a robust learning rule for $L^2(\XX)$, we argue that this hypothesis class is not appropriate for the analysis of the adversarial examples phenomenon in ANNs. Since it is not a normal hypothesis class, the hypotheses in $L^2(\XX)$ are not clustered together and it does not model the transferability of the adversarial examples between ANNs with different activation functions.

\section{The space of holomorphic hypotheses}
\label{sec:holo}
In \secref{sec:pre} we introduced $\ann{L^2(\Omega)}{\sigma(\XX)}$ and generalized the SVC learning rule to this hypothesis class. However, we could infer that $\ann{L^2(\Omega)}{\sigma(\XX)}$ is not a normal hypothesis class in general. In order to overcome this obstacle, we will turn to a special subset of $L^2(\XX)$.

\begin{definition}[the Bergman space]
Consider a compact and simply connected domain set $\XX\subset\mathbb{C}^d$. The Bergman space $A^2(\XX)\subset L^2(\XX)$ is a reproducing kernel Hilbert space defined as
\begin{equation}
A^2(\XX)=
\{f\in\mathcal{O}(\XX)\,|\,\Big(\int_\XX|f(z)|^2dV(z)\Big)^\frac{1}{2}<\infty\}.
\end{equation}
\end{definition}

\begin{theorem}
$A^2(\XX)$ and $\partial A^2(\XX)$ are normal hypothesis classes.
\label{thm:transfer}
\end{theorem}

\Thmref{thm:transfer} is exactly the result that we are looking for. $\mathcal{O}(\XX)$ is the space of holomorphic functions on $\mathcal{X}$. There are different equivalent ways to characterize holomorphic functions. The holomorphic functions are the solutions to the homogeneous Cauchy-Riemann equations, or its counterpart in higher dimensions $\overline{\partial}$ (del-bar) equations. Equivalently, the holomorphic functions are the functions that are complex differentiable in each dimension of $z\in\XX$. A third characterization of $\mathcal{O}(\XX)$ is that these functions are complex analytic and have a power series representation. $\mathcal{O}(\XX)$ is also special because, unlike real analytic functions, it is closed under uniform convergence. In one dimension, holomorphic functions are also known as the conformal maps. However, it could be shown that holomorphic functions can only be defined over the field of complex numbers. Nevertheless, we argue that the simplicity of analysis in $A^2(\XX)$ relative to other normal hypothesis classes, and its unique set of properties would justify the switch to complex numbers, as long as it results into methods that could be translated to a counterpart for real numbers. Consequently, we have to extend the SVC learning rule to accommodate for complex-valued hypotheses. We have replaced $x$ with $z$ to symbolize the transition from the real to the complex number system.

A domain $\XX$ that allows for the definition of holomorphic functions is called a domain of holomorphy. Fortunately, most of the domains that we would normally face in applications are domains of holomorphy. A notable subset of domains of holomorphy are convex domains. We have provided a short summary of the topic in appendix \ref{sec:domains}, with some examples on how to encode real data with complex numbers. \cite{krantz2001function} is our main reference for the definitions and notation in function theory of several complex variables.

\begin{definition}[complex-valued classifier]
A complex-valued classifier $f:\Omega\rightarrow\mathbb{C}$ is a function $f(z)=u(z)+iv(z)$ in which the real part $\Re[f(z)]=u(z)=0$ encodes the geometrical position of the decision boundary and the imaginary part $\Im[f(z)]=v(z)=0$ regresses through the geometrical position of the training samples. 
\end{definition}

\begin{definition}[the Bergman kernel]
The Bergman kernel $K_\XX(z,\zeta)$ of a compact and simply connected domain $\XX$ is the unique function with the reproducing property
\begin{equation}
f(z)=\int_\XX f(\zeta)K_\XX(z,\zeta)dV(\zeta),\quad \forall f\in A^2(\XX).
\end{equation}
\end{definition}

The reproducing property of the Bergman kernel of $\XX$ in conjunction with the fact that the optimal Bayes classifier achieves the minimum of the 0-1 loss function provides the means to define the infinite sample limit of any learning rule on $A^2(\Omega)$ that is minimizing the complex 0-1 loss function,
\begin{equation}
    \ell_\mathbb{C}(t,z,f)=\ell_{0-1}(t,z,\Re[f])+\ell_{MSE}(0,z,\Im[f]),
\end{equation}
independently from the details of the implementation or the training process.

\begin{definition}[holomorphic optimal Bayes classifier]
\label{defn:hobc}
The holomorphic optimal Bayes classifier is the orthogonal projection of the optimal Bayes classifier $f_\mathcal{D}:\XX\rightarrow\mathbb{R}$ into $A^2(\XX)$,
\begin{equation}
o_\mathcal{D}(z)=\int_\XX f_\mathcal{D}(\zeta)K_\XX(z,\zeta)\,dV(\zeta),
\end{equation}
\end{definition}

Following \cite{shamir2021dimpled} and \cite{tanay2016boundary}, we are interested to know how would $\SSS(f)=\{z\in\XX\,|\,\Im[f(z)]=0\}$ and $\CC(f)=\{z\in\XX\,|\,\Re[f(z)]=0\}$ interact with the geometrical position of the training samples. Geometrical properties of holomorphic functions would enable us to infer the geometrical relation between the manifold of the samples and the decision boundary by studying the image of the manifold in the range space of the hypothesis. This property of the holomorphic hypotheses would prove to be key to finding a robust learning rule for $A^2(\XX)$.

\begin{definition}[complex-valued SVC learning rule]
Consider a dataset $S=\seq{(z_n,t_n)\in\XX\times\{-1,+1\}}{n=1}{N}$ and a hypothesis $f \in \ann{A^2(\Omega)}{\sigma(\XX)}$. The complex-valued SVC learner solves the following program,
\begin{argmini}
{h,\xi}{\frac{1}{2}\|h\|_{A^2(\Omega)}^2+C\sum_{n=1}^N\xi_n}
{\label{opt:lmm3}}{}
\addConstraint{\xi_n}{\geq 0}
\addConstraint{t_n\Re[f(z_n)]}{\geq 1 - \xi_n,}
\addConstraint{\xi_n}{\geq \Im[f(z_n)])}
\addConstraint{\Im[f(z_n)]}{\geq -\xi_n}
\end{argmini}
in which $\seq{\xi_n}{n=1}{N}$ are slack variables introduced to allow for soft margins.
\end{definition}

\section{Robust classification for holomorphic hypotheses}
\label{sc:robust}
In this section, we will develop a robust learning rule with respect to $A^2(\XX)$. To do so, we first examine a toy problem to get an intuition for how adversarial examples occur in holomorphic hypotheses, and then continue to introduce the proposed robust learning rule.

To start, we will try to classify the unit disk $\mathbb{D}$ into distinct halves in which
\begin{equation}
    t(z)=\mathrm{sign}(\Re[z]) \quad z\in\mathbb{D},
\end{equation}
is the labeling function. We will choose our training samples to be the set
\begin{equation}
S_n=\Big\{\big(z,t(z)\big)\,|\,z=e^{i\frac{k}{n}2\pi}\quad k=0,\cdots,n-1\Big\}.    
\end{equation}
We will use the orthonormal polynomial basis of the unit disk as features,
\begin{equation}
    \varphi_k(z)=\sqrt{\frac{k+1}{\pi}}z^k.
\end{equation}

We will visualize a holomorphic function in three ways. First, we use a domain coloring technique and graph the hypotheses on $\mathbb{D}$. The hue of a color represent the angle, and the saturation represent the magnitude of a complex number. We will also plot the contours of the real and imaginary parts of the hypotheses in the same graph using white and black lines respectively. Second, we will graph the real and the imaginary parts of the hypotheses on the unit circle $\mathbb{T}=\partial\mathbb{D}$. $\mathbb{T}$ is the set in which all the training points of $S_n$ are sampled from. Third, we will graph the image of $\mathbb{T}$ in the range space of the hypotheses. In the range space of a complex-valued hypothesis, the decision boundary $\CC(f)$ is represented by the imaginary axis, and the real axis represent the curve that has regressed through the training points $\SSS(f)$.

\begin{figure*}
     \centering
     \begin{subfigure}[b]{\textwidth}
         \centering
         \includegraphics[width=\textwidth]{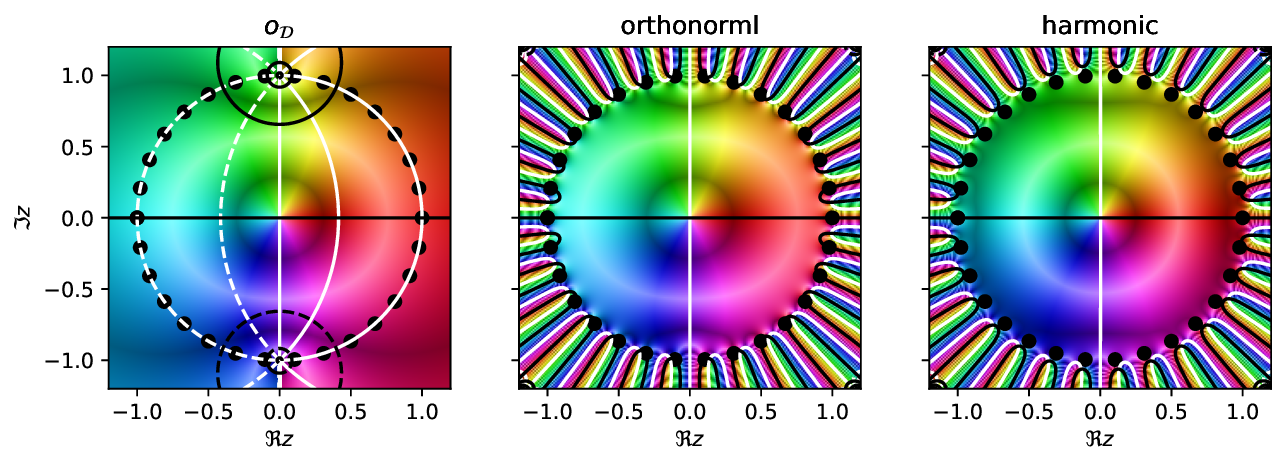}
         \caption{graph of the complex-valued hypotheses on the unit disk $\mathbb{D}$.}
         \label{fig:optimal disk}
     \end{subfigure}
     \hfill
     \begin{subfigure}[b]{\textwidth}
         \centering
         \includegraphics[width=\textwidth]{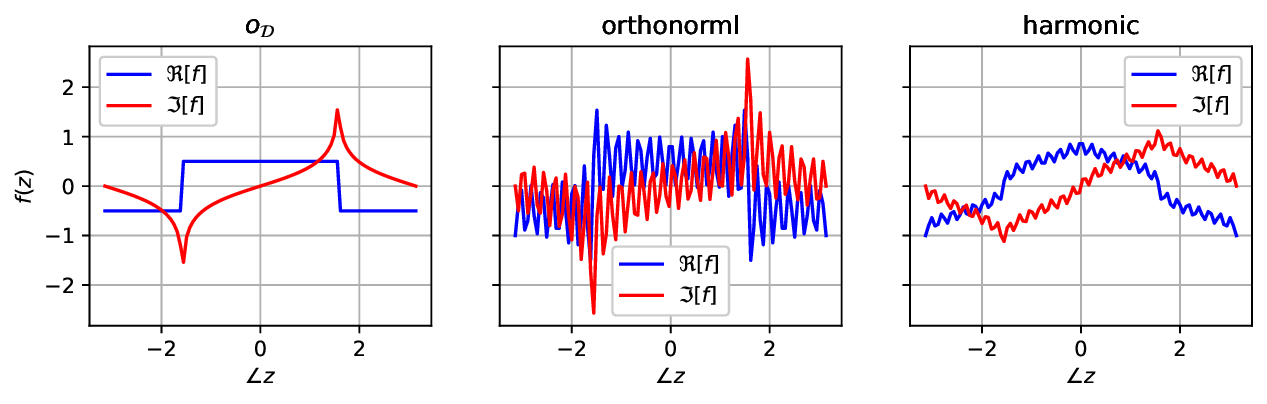}
         \caption{graph of the complex-valued hypotheses on the unit circle $\partial\mathbb{D}$.}
         \label{fig:optimal circle}
     \end{subfigure}
     \hfill
     \begin{subfigure}[b]{\textwidth}
         \centering
         \includegraphics[width=\textwidth]{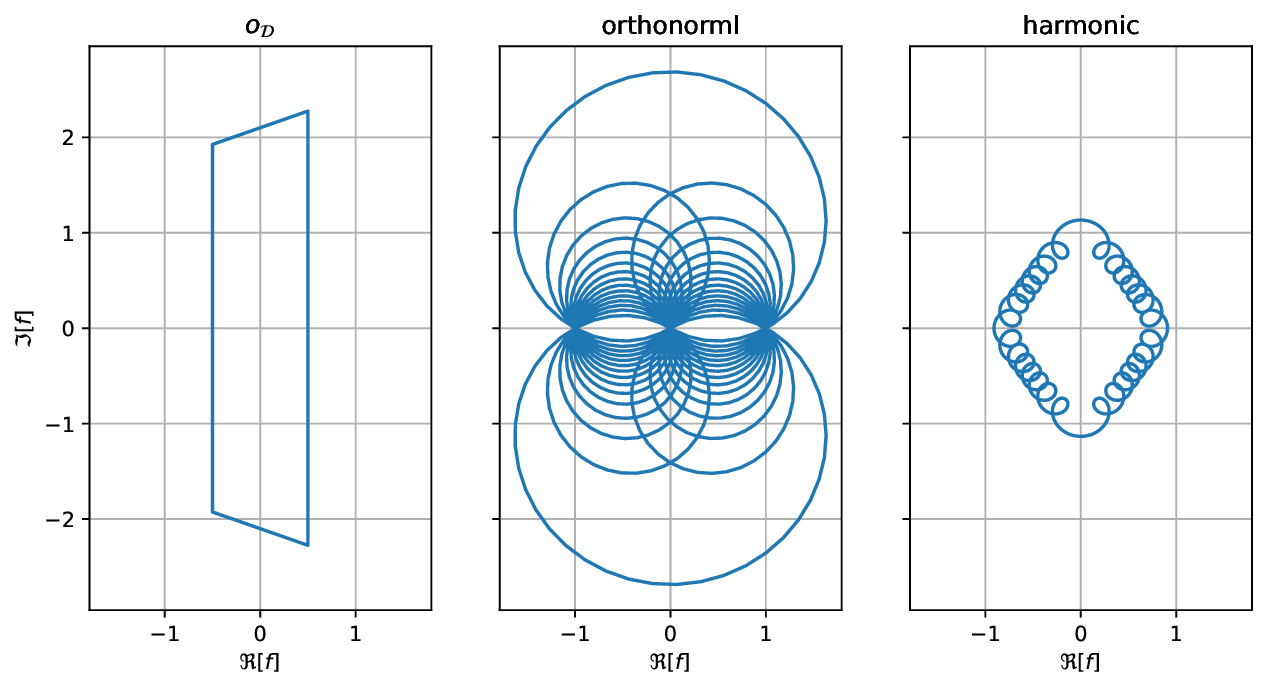}
         \caption{image of $\mathcal{X}=\partial\mathbb{D}$ under the complex-valued hypotheses.}
         \label{fig:optimal strip}
     \end{subfigure}
    \caption{Visualizations of $o_\mathcal{D}$ and the optimal hypotheses of the hinge loss function for $S_{30}$. We can see that even though the hypotheses are optimizing a different loss function than the 0-1 loss, they show certain similarities with $o_\mathcal{D}$.}
    \label{fig:optimal}
\end{figure*}

We have visualized the analogue of holomorphic optimal Bayes classifier $o_\mathcal{D}$ in the left column of \figref{fig:optimal}. However, we have made use of the Szeg\H{o} kernel to project the labeling function $t$ instead of the Bergman projection of $t$. The middle column of \figref{fig:optimal} depicts the output of applying the complex-valued SVC learning rule to $S_{30}$ using the orthonormal basis of $A^2(\mathbb{D})$ as features. As we have demonstrated in \figref{fig:optimal}, the solution to \optref{opt:lmm3} is not robust with respect to $A^2(\mathbb{D})$. Nevertheless, due to the normality of the holomorphic functions, we can see that the learned hypothesis resembles $o_\mathcal{D}$, even though they are not learned through the same learning rule, or make use of the same hypothesis class for that matter.

Looking at \figref{fig:optimal}, we can see that $o_\mathcal{D}$ has two logarithmic branch points on $i$ and $-i$. With this observation in mind, it is no wonder that approximating $o_\mathcal{D}$ is troublesome for learning rules. Furthermore, we see that the natural boundary of the nonrobust hypothesis has advanced inside of $\mathbb{D}$. This advancement has caused the learners output to be nonrobust in a neighborhood of $\mathbb{T}$. This observation is common to nonrobust learning rules with respect to $A^2(\XX)$.

\begin{theorem}
\label{thm:paradox}
Let $f\in A^2(\XX\subset\mathbb{C}^d)$ be the output of a nonrobust learning rule, then $f$ is robust on a dense open subset of $\XX$. 
\end{theorem}

\Thmref{thm:paradox} shows that even though our choice of $A^2(\XX)$  as a replacement for $L^2(\XX)$ was mainly derived by the transferability of adversarial examples, it also correctly predicts the apparent paradox in the existence of the adversarial examples in ANNs, where the nonrobust hypothesis appears to be robust almost everywhere.

\Figref{fig:optimal strip} graphs the image of $\mathbb{T}$ in the range space of the hypotheses $f(\mathbb{T})$. Due to the singularities of the holomorphic projection of $t$, $o_\mathcal{D}(\mathbb{T})$ passes through the point at the infinity. Consequently, $o_\mathcal{D}(\mathbb{T})$ consists of two parallel lines; the trapezoidal shape in the figure is caused by the fact that we cannot in practice reach the point at the infinity. Looking at the nonrobust hypothesis in \Figref{fig:optimal strip}, we can see that the image of $\mathbb{T}$ passes through the decision boundary quite a few times. The figure shows that $f(\mathbb{T})$ of the nonrobust hypothesis is longer than necessary. In other words, it is possible to find a holomorphic hypothesis that achieves the same loss on the training set with $f(\mathbb{T})$ not passing through the decision boundary so many times. We can repeat the same argument for any other curve inside $\mathbb{D}$. Thus, we argue that if the output $f$ of $\AAA$ with respect to $A^2(\XX)$ minimizes the area covered by the image of $\XX$ under $f$, then it is robust.

\begin{theorem}
\label{thm:dirichlet}
A learning rule $\AAA$ with respect to $A^2(\XX)$ that minimizes the Dirichlet energy of its output,
\begin{equation}
    E[f]=\int_\XX\|\nabla f(z)\|^2\,dV(z),
\end{equation}
is robust.
\end{theorem}

\begin{definition}[robust (complex-valued) SVC learning rule]
The robust (complex-valued) SVC learning rule is the same as the (complex-valued) SVC learning rule, but minimizes $E[h]$ instead of $\|h\|_{\HH(\XX)}^2$.
\end{definition}

\begin{definition}[harmonic features]
A set of non-constant features $\{\varphi_k\}_1^\infty$ is harmonic if
\begin{equation}
    \int_\XX\nabla\varphi_j(z)\cdot\nabla\varphi_k(z)\,dV(z)=
    \begin{cases}
    1& \quad j=k\\
    0& \quad j\neq k
    \end{cases},
\end{equation}
\end{definition}

Given the tuning matrix
\begin{equation}
    \Sigma_{jk}=\int_\XX \nabla\varphi_j(z)\cdot\nabla\varphi_k(z)\,dV(z),
\end{equation}
we can transform any set of features to its harmonic counterpart,
\begin{equation}
    \varphi^*=\Sigma^{-\frac{1}{2}}\varphi.
\end{equation}
We emphasize that $\Sigma$ is positive definite by definition. Thus, a unique square root of $\Sigma$ exists and it is invertible.

\begin{theorem}
\label{thm:ell2}
$\ell^2$ regularization of a harmonic hypothesis minimizes its Dirichlet energy.
\end{theorem}

We have repeated the toy experiment using the harmonic basis functions of $A^2(\XX)$, and reported the results in the right column of \figref{fig:optimal}. We can see that the harmonic hypothesis is robust as expected. Thus, we have succeeded in finding a robust learning rule for $A^2(\XX)$. The interested reader can find more examples and experiments in the supplementary material of the paper.

\section{Robust training of ANNs}
\label{sec:ann}
In this section, we will apply the results of \secref{sc:robust} to ANNs in a limited manner to demonstrate the feasibility and applicability of our approach. First, we put the proposed framework to use and describe how adversarial examples occur in ANNs.
\begin{figure}
    \centering
    \includegraphics[width=0.45\textwidth]{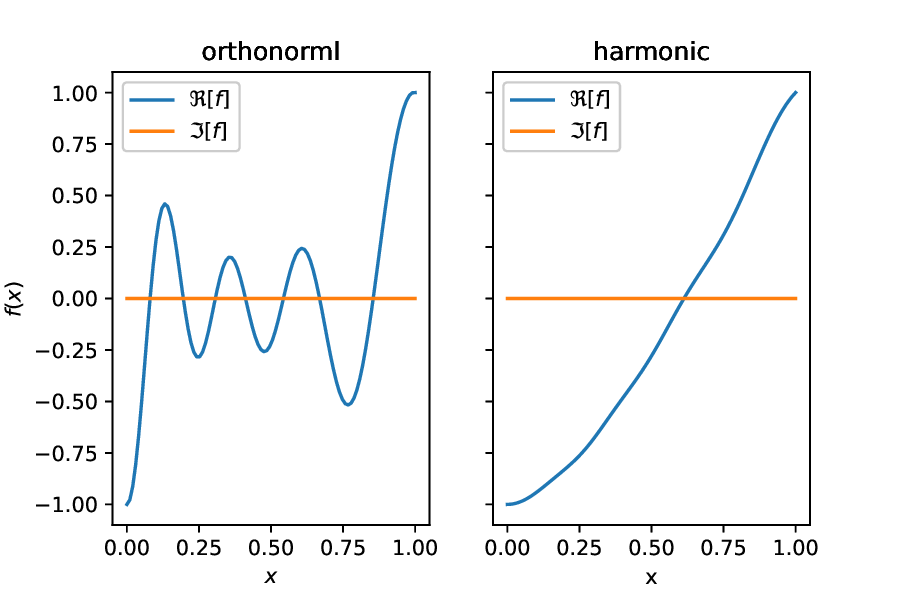}
    \caption{The output of the robust (right) and nonrobust (left) SVC learning rules for $\ann{A^2(\mathbb{D})}{\mathrm{ReLU}([0,1])}$.}
    \label{fig:ann real}
\end{figure}

\begin{figure}
    \centering
    \includegraphics[width=0.45\textwidth]{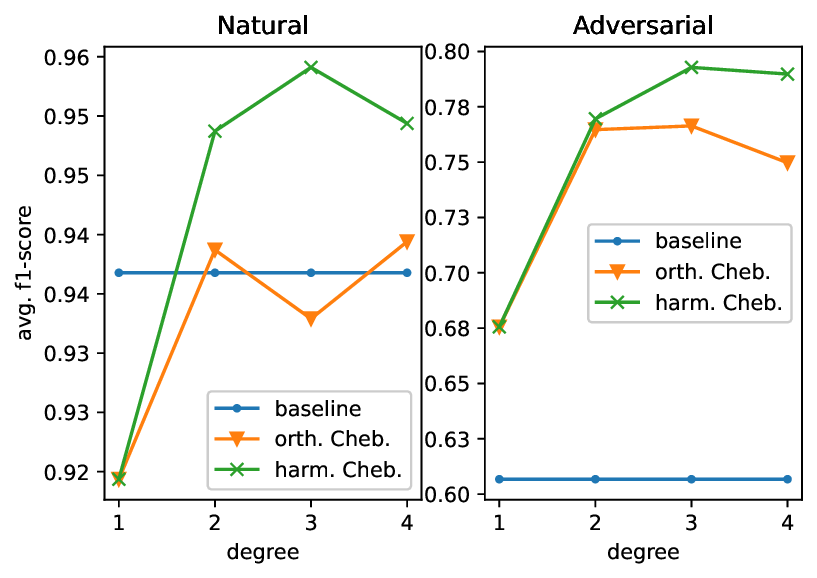}
    \caption{The result of performing a one-step $\ell^2$ normalized gradient-based black-box attack on harmonic and orthonormal Chebyshev polynomials for the UCI digits dataset. The baseline is a $40 \times 30 \times 20$ fully-connected MLP that is trained on the same dataset}
    \label{fig:digits}
\end{figure}

We already know that the orthonormal basis for $A^2(\Omega)$ would not be robust on the boundary of $\Omega$. In other words, the learning rule would fail to find the robust coefficients of neurons on the boundary of $\Omega$. When we use the harmonic bases on the other hand, the region of convergence of $h(\omega)$ would cover $\partial\Omega$ and the output of the learning rule would be robust. We have computed the first 30 orthonormal and the harmonic projections $\seq{\psi_n=\int_\Omega \varphi_n(\omega)\sigma(x;\omega)\,dV(\omega)}{n=1}{\infty}$ of the activation function $\mathrm{ReLU}\big(\Re[\omega]x+\Im[\omega]\big)$ for $x\in[0,1]$, and applied the complex-valued SVC learning rule of $\ann{A^2(\mathbb{D})}{\mathrm{ReLU}(\XX)}$ to the toy example of \cite{9412367}. We have reported the results of the experiment in \figref{fig:ann real}. It is clear from the figure that the harmonic ANN is very robust. However, it is also clear that the output of the learner has not maximized the margin. The problem persists for $\ann{L^2(\mathbb{D})}{\sigma(\XX)}$ as well. We believe that the most likely reason is that we have not chosen the best domain set $\Omega$ to parameterize the neurons and as a result the learners output is biased. We believe that a satisfying study of the parameter domain sets $\Omega$ of generalized ANNs is out of the scope of the current paper, and we leave it for future studies.

To showcase the applicability of the framework in a real-world scenario, we use a MLP classifier with ReLU activation as a surrogate and attack two polynomial hypotheses that are trained on a subset of the UCI digits dataset \citep{Dua:2019}. We have reported the results in \figref{fig:digits}. The results show that the proposed robust learning rule is effective in mitigating the effects of the phenomenon, and supports the proposition that ANNs and polynomials are different representations of the same normal hypothesis class.

\begin{theorem}
\label{thm:pde}
The output of the robust SVC learning rule with respect to $\ann{L^2(\Omega)}{\sigma(\XX)}$ satisfies the Poisson partial differential equation (PDE)
\begin{equation}
    -\Delta h(\omega)=\sum_{n=1}^N\lambda_nt_ns(x_n;\omega),
\end{equation}
in which $\Delta$ is the Laplace operator.
\end{theorem}

\Thmref{thm:pde} shows that training a robust ANN is the same as solving a PDE. This observation suggests that robust classifiers are a normal family of functions. The fundamental solution $\Phi$ of the Laplace operator could be used to find $h$ as a function of $\seq{\lambda_n}{n=1}{N}$,
\begin{equation}
    h(\omega)=\sum_{n=1}^N\lambda_nt_n\int_\Omega\Phi(\omega-w)s(x_n;w)\,dV(w).
\end{equation}

\begin{definition}[harmonic activation function]
A family of activation functions $\mathfrak{S}(\XX)$ is harmonic if they satisfy the Helmholtz equation with the natural Dirichlet or Neumann boundary condition
\begin{equation}
    -\Delta_\omega\mathfrak{s}(x;\omega)=\mathfrak{s}(x;\omega)\quad \mathfrak{s}\in\mathfrak{S}(\XX).
\end{equation}
\end{definition}

\begin{theorem}
\label{thm:final}$L^2$ regularization of $h(\omega)$ of a harmonic generalized ANN $f\in\ann{L^2(\Omega)}{\mathfrak{S}(\XX)}$ is the same as minimizing its Dirichlet energy.
\end{theorem}

\Thmref{thm:final} is the final result of this paper. We have managed to find a robust learning rule for a generalized notion of ANNs. Solving PDEs is an involved and complicated process, and we will not attempt to solve the derived PDEs in this paper. Nevertheless, both PDEs are very well-known and well-studied, and various methods and techniques are available in the literature which focuses on solving these PDEs.

\section{Conclusion}
In this paper we introduced a general framework for training robust ANN classifiers. We proposed a formal definition for robustness and transfer in learning theoretic terms, and describe how adversarial examples might emerge from the pointwise convergence of the trained hypothesis to the infinite sample limit of the learning rule. Since our proposal assumes that the convergence is pointwise, our proposal needs a separate explanation for the transfer of adversarial examples between hypotheses under pointwise convergence. To this end, we propose the normal hypothesis classes, and define these classes by adapting the definition of a normal family of functions in analysis. We also introduce generalized ANNs as an abstraction of ANNs that is easier to analyze in terms of convergence.

Next, we showed that under the proposed definitions, learning rules for ANNs that are converging to a hypothesis in $L^2(\XX)$ does not explain the observation of transfer between different architectures of ANNs. As an alternative, we propose that transfer in ANNs is better modeled by functions in $\mathcal{O}(\XX)$. $\mathcal{O}(\XX)$ is the set of holomorphic functions on $\XX$. We provide the necessary definitions to use $\mathcal{O}(\XX)$ as a hypothesis class, and take the first steps to enable the use of powerful tools of complex analysis in the study of the phenomenon.

Through holomorphicity, our framework provides a geometrical interpretation for the adversarial examples phenomenon. We conclude that a binary classifier with minimal Dirichlet energy is robust. In other words, replacing the $L^2$ regularization term in the loss function with the Dirichlet energy of the hypothesis should result in a robust classifier. Minimizing the Dirichlet energy might not be tractable in gradient descent as we need to compute the derivative of the Dirichlet energy to do so. To circumvent this problem, we introduce the harmonic features and activation functions. In summary, we first construct features or activation functions that satisfy a condition, and then show that $L^2$ regularization of these hypothesis classes is the same as minimizing their Dirichlet energy. Consequently, minimizing the Dirichlet energy for these classes could be as efficient as $\ell^2$ regularization of their parameters.

To the best of our knowledge, we have provided the first method that makes use of calculus of variations and differential equations to tackle the challenge of robust training of ANNs. Nevertheless, the analysis proved that training robust ANNs classifiers is not trivial, and we need to either implicitly or explicitly solve an intermediate PDE. There are multiple methods of doing so available in the relevant literature. As a result, we have left the actual implementation of the proposed method to future, when an in depth understanding of the appropriate parameter domain sets and harmonic activation functions in ANNs is achieved.

\bibliography{main.bib}

\section*{Checklist}

The checklist follows the references.  Please
read the checklist guidelines carefully for information on how to answer these
questions.  For each question, change the default \answerTODO{} to \answerYes{},
\answerNo{}, or \answerNA{}.  You are strongly encouraged to include a {\bf
justification to your answer}, either by referencing the appropriate section of
your paper or providing a brief inline description.  For example:
\begin{itemize}
  \item Did you include the license to the code and datasets? \answerYes{}
  \item Did you include the license to the code and datasets? \answerNo{The code and the data are proprietary.}
  \item Did you include the license to the code and datasets? \answerNA{}
\end{itemize}
Please do not modify the questions and only use the provided macros for your
answers.  Note that the Checklist section does not count towards the page
limit.  In your paper, please delete this instructions block and only keep the
Checklist section heading above along with the questions/answers below.

\begin{enumerate}

\item For all authors...
\begin{enumerate}
  \item Do the main claims made in the abstract and introduction accurately reflect the paper's contributions and scope?
    \answerYes{}
  \item Did you describe the limitations of your work?
    \answerYes{}
  \item Did you discuss any potential negative societal impacts of your work?
    \answerNA{}
  \item Have you read the ethics review guidelines and ensured that your paper conforms to them?
    \answerYes{}
\end{enumerate}

\item If you are including theoretical results...
\begin{enumerate}
  \item Did you state the full set of assumptions of all theoretical results?
    \answerYes{}
        \item Did you include complete proofs of all theoretical results?
    \answerYes{}
\end{enumerate}

\item If you ran experiments...
\begin{enumerate}
  \item Did you include the code, data, and instructions needed to reproduce the main experimental results (either in the supplemental material or as a URL)?
    \answerYes{}
  \item Did you specify all the training details (e.g., data splits, hyperparameters, how they were chosen)?
    \answerNA{}
        \item Did you report error bars (e.g., with respect to the random seed after running experiments multiple times)?
    \answerNA{}
        \item Did you include the total amount of compute and the type of resources used (e.g., type of GPUs, internal cluster, or cloud provider)?
    \answerNA{}
\end{enumerate}

\item If you are using existing assets (e.g., code, data, models) or curating/releasing new assets...
\begin{enumerate}
  \item If your work uses existing assets, did you cite the creators?
    \answerNA{}
  \item Did you mention the license of the assets?
    \answerNA{}
  \item Did you include any new assets either in the supplemental material or as a URL?
    \answerNA{}
  \item Did you discuss whether and how consent was obtained from people whose data you're using/curating?
    \answerNA{}
  \item Did you discuss whether the data you are using/curating contains personally identifiable information or offensive content?
    \answerNA{}
\end{enumerate}

\item If you used crowdsourcing or conducted research with human subjects...
\begin{enumerate}
  \item Did you include the full text of instructions given to participants and screenshots, if applicable?
    \answerNA{}
  \item Did you describe any potential participant risks, with links to Institutional Review Board (IRB) approvals, if applicable?
    \answerNA{}
  \item Did you include the estimated hourly wage paid to participants and the total amount spent on participant compensation?
    \answerNA{}
\end{enumerate}

\end{enumerate}

\newpage
\appendix

\section{Comparison with the literature}
The main point of distinction between the proposed framework and the competing proposals is in the use of complex analysis to simplify the analysis. Even though analytic properties of the hypotheses has been exploited before in the literature to derive various methods of defence and attack, e.g. \cite{10.1145/3522749.3523078}, the use of holomorphic functions and complex analysis in machine learning have been sparse \cite{10.1145/3208976.3208989,10.1145/2688073.2688076,Das_Sarma_2019}. We believe the reason to be that applying complex analysis to machine learning is an interdisciplinary effort by nature.

\cite{goodfellow2014explaining} were the first to show that nonrobust ANNs are weak to analytic attacks, and attributed the phenomenon to something that they called linearity. \cite{tanay2016boundary} refuted this claim by showing that linear classifiers could be robust. We answer the apparent paradox between linear and nonlinear nature of the phenomenon by proposing that both positions are in some sense correct, and that the phenomenon would be better described by the analytic properties of the robust hypothesis.

\cite{DBLP:conf/nips/HeinA17} proposed a certificate of robustness for differentiable classifiers. The same certificate could be used for analytic functions. The certificate might be improved by considering the Taylor series expansion of the classifier around a test sample. The Abel-Ruffini theorem might result in some complications in this process.

From a geometric point of view, our proposal is very well aligned with the dimpled manifold model of \cite{shamir2021dimpled}. Regarding the boundary tilting perspective of \cite{tanay2016boundary}, it seems that the nonrobust classifier in \figref{fig:optimal strip} has managed to cross the decision boundary at right angles, and there seems to be exceptions to the boundary tilting perspective in nonlinear cases. \cite{arxiv.2103.00778} hypothesises that the robust classifier would maximize the margin. We did not make an attempt to analyze the learning rule from the perspective of maximum margin classification. Nonetheless, the peculiar shape of the image of $\mathbb{T}$ under the robust hypothesis in \figref{fig:optimal strip} suggests that the proposed robust SVC learning rule also maximizes the margin in some sense.  

Our model for the transfer of adversarial examples is mostly similar to the proposal of \cite{goodfellow2014explaining}, in which the reason for the transfer of adversarial examples are deemed to be that the models converge to the optimal linear hypothesis. \cite{DBLP:journals/corr/PapernotMG16} shows that existing machine learning approaches are in general vulnerable to systematic black-box attacks regardless of their structure, showing that transferable adversarial examples are common in machine learning models. Here, we have defined the holomorphic optimal Bayes classifier and given a formal definition of how transfer occurs. \cite{ilyas2019adversarial} moves the blame to hidden patterns in the input, which our proposal does not align with. \cite{8953700} relates the transfer of the adversarial examples to the learned features as well.

There are multiple instances of the use of gradient information through out the literature \citep{ross2018improving,arxiv.2103.00778}. The main contrast between our proposal is that we manage to relate gradient regularization with the geometry of $f(\XX)$ and further reveal how gradient regularization is related with the Dirichlet energy of the hypothesis.

\section{Proof of the theorems}
\subsection{\Thmref{thm:robustness}}
\begin{proof}
Consider a nonuniform learner $\AAA$ with respect to $\HH(\XX)$. Suppose that we have a sequence $\seq{S_n}{n=1}{\infty}$ of dense training sets $S_n\subseteq\XX$ in which
\begin{equation}
    \lim_{n\rightarrow\infty}\,S_n=\XX.
\end{equation}
Then, $\seq{f_n=\AAA(S_n)}{n=1}{\infty}$ converges compactly to $f=\AAA(\XX)$. By Cauchy's criterion for uniform convergence, for any $\epsilon>0$ a natural number $N<\infty$ exists in which
\begin{equation}
    \sup_{x\in\XX} |f_m(x)-f_n(x)|\leq\epsilon \quad m,n\geq N.
\end{equation}

Now suppose that an adversarial test point $x\notin S_n$ for $f_n$ exists. We know that for some $m>n$ it is true that $x\in S_m$. Thus,
\begin{equation}
    \sup_{x\in\XX} |f_m(x)-f_n(x)|>\epsilon,
\end{equation}
which contradicts our assumption. Consequently, it must be true that no adversarial examples for $f_n$ exists.
\end{proof}

\subsection{\Lemref{lem:subset}}
\begin{proof}
Since $\sigma(\XX)$ has the universal approximation property, then for all $\alpha\in\mathbb{N}$ an $\omega\in\Omega$ exists where
\begin{equation}
    \psi_\alpha(\omega)=\int_\XX\varphi_\alpha(x)s(x;\omega)\,dV(\XX)\neq0.
\end{equation}
$\seq{\varphi_\alpha}{\alpha=0}{\infty}$ is the complete orthonormal basis for $\HH(\XX)$. It is easy to check that any $f\in\ann{\HH(\Omega)}{\sigma(\XX)}$ has a representation
\begin{equation}
    f(x)=\sum_{\alpha=0}^{\infty}a_\alpha\psi'_\alpha(x),
\end{equation}
in which the projected features $\seq{\psi'_\alpha}{\alpha=0}{\infty}$ are computed by 
\begin{equation}
    \psi_\alpha'(x)=\int_\Omega\varphi'_\alpha(\omega)s(x;\omega)\,dV(\omega),
\end{equation}
where $\seq{\varphi'_\alpha}{\alpha=0}{\infty}$ is the complete orthonormal basis for $\HH(\Omega)$.

Consider the inner product of $\psi'_\alpha$ with $\varphi_\beta$,
\begin{align}
    \psi'_\alpha\cdot\varphi_\beta&=\int_\XX\psi'_\alpha(x)\varphi_\beta(x)\,dVx,\\
    &=\int_\XX\int_\Omega\varphi_\beta(x)\varphi'_\alpha(\omega)s(x;\omega)\,dV(\omega)dV(x),\\
    &=\int_\Omega\varphi'_\alpha(\omega)\int_\XX\varphi_\beta(x)s(x;\omega)\,dV(x)dV(\omega),\\
    &=\int_\Omega\varphi'_\alpha(\omega)\psi_\beta(\omega)\,dV(\omega),\\
    &=\varphi_\alpha'\cdot\psi_\beta.
\end{align}
Suppose that some $\varphi_\beta$ is orthogonal to all $\psi'_\alpha$. Then, we can deduce that $\psi_\beta$ is orthogonal to all $\varphi'_\alpha$. Given that $\seq{\varphi'_\alpha}{\alpha=0}{\infty}$ is a complete basis for $\HH(\Omega)$, this would lead to a contradiction $\psi_\beta=0$. Thus, $\HH(\XX)\subseteq\ann{\HH(\Omega)}{\sigma(\XX)}$.

To see that $\ann{\HH(\Omega)}{\sigma(\XX)}\subseteq\HH(\XX)$ is also true, we have to consider a few cases. Since $\sigma(\XX)$ has the universal approximation property, we know that the dimension of $\Omega$ is greater than or equal to the dimension of $\XX$.

First, suppose that the dimension of $\Omega$ is greater than $\XX$. This would mean that it for each $\varphi_\beta$ one or more $\psi'_\alpha$ exists. Given that $\ann{\HH(\Omega)}{\sigma(\XX)}\subseteq L^2(\XX)$, it must be true that $\seq{\psi'_\alpha}{\alpha=0}{\infty}$ is redundant. Thus, we can find a sequence $\seq{\psi"_\beta}{\beta=0}{\infty}$ which has the same index as $\seq{\varphi'_\beta}{\beta=0}{\infty}$ by performing linear transformations on $\seq{\psi'_\alpha}{\alpha=0}{\infty}$. As a result, this case could be transformed to an equivalent situation in which the dimensions of $\Omega$ and $\XX$ are equal.

Now, suppose that the the two domain sets have the same dimension. Then, a bijection between $x\in\XX$ and $\omega\in\Omega$ exists. Otherwise, $\ann{L^2(\Omega)}{\sigma(\XX)}$ would not have enough degrees of freedom to universally approximate $L^2(\XX)$. Consequently, a bijection between $\HH(\XX)$ and $\HH(\Omega)$ exists. A bijection between $\HH(\Omega)$ and $\ann{\HH(\Omega)}{\sigma(\XX)}$ exists by definition as well. In other words, $|\HH(\XX)|=|\HH(\Omega)|=|\ann{\HH(\Omega)}{\sigma(\XX)}|$. However, we just showed $\HH(\XX)\subseteq\ann{\HH(\Omega)}{\sigma(\XX)}$. Thus, it must be true that $\HH(\XX)=\ann{\HH(\Omega)}{\sigma(\XX)}$.

If on the other hand $\sigma(\XX)$ does not have the universal approximation property, then it is possible that $\psi_\alpha=0$ for none or all $\alpha$. 
\end{proof}

\subsection{\Thmref{thm:non learnability}}
\begin{proof}
$L^2(\XX)$ would be nonuniform learnable if and only if it is a countable union of PAC learnable hypothesis classes $\HH_\alpha(\XX)$. Since $L^2(\XX)$ is a Hilbert space, it has a orthonormal basis $\seq{\varphi_\alpha}{\alpha=0}{\infty}$. Then, we can choose
\begin{equation}
    \HH_\alpha(\XX)=\big\{\sum_{k=0}^{\alpha}a_k\varphi_k(x)|a_k\in\mathbb{R}\big\},
\end{equation}
and the theorem would follow.
\end{proof}

\subsection{\Thmref{thm:svc consistency}}
\begin{proof}
The Lagrangian of \optref{opt:lmm} is
\begin{equation}
    \mathcal{L}=\frac{1}{2}\int_\Omega|h(\omega)|^2\,dV(\omega)+\sum_{n=1}^N\lambda_n\big(1-t_n\int_\Omega h(\omega)s(x_n;\omega)\,dV(\omega)\big).
\end{equation}
We can rearrange the Lagrangian into
\begin{align}
    \mathcal{L}&=\int_\Omega\frac{1}{2}|h(\omega)|^2-\sum_{n=1}^N\lambda_nt_nh(\omega)s(x_n;\omega)\,dV(\omega) + \sum_{n=1}^N\lambda_n,\\
    &=\int_\Omega L(\omega,h,\nabla h)\,dV(\omega) + \sum_{n=1}^N\lambda_n.
\end{align}
By the Euler-Lagrange equations for a function of several variables, $h^*$ must satisfy the following PDE
\begin{equation}
    \frac{\delta L}{\delta h}-\sum_{j=1}^d\frac{\partial}{\partial x_j}\frac{\delta  L}{\delta h_j}=0,
\end{equation}
in which $h_j=\frac{\partial h}{\partial x_j}$, and $\frac{\delta}{\delta h}$ is the usual differentiation with the exception that it treats $h$ like a symbolic variable. Thus,
\begin{equation}
    h^*(x)=\sum_{n=1}^N\lambda_nt_ns(x_n;\omega).
\end{equation}
Consider the Dirac's delta
\begin{equation}
    \delta(x;\omega)=
\begin{cases}
\infty \quad x=\omega\\
0 \quad \mathrm{otherwise}
\end{cases}.
\end{equation}
We can see that the output of the SVC learning rule for $\ann{L^2(\XX)}{\mathrm{Dirac}(\XX)}=L^2(\XX)$ would be the same as the memorising learning rule, which is the text book example of a consistent learner that is not nonuniform. Thus, the SVC learning rule is not a nonuniform learner with respect to $\ann{L^2(\Omega)}{\sigma(\XX)}$ in general.
\end{proof}

\subsection{\Propref{prop:l2 transfer}}
\begin{proof}
Consider $\ann{L^2(\XX)}{\mathrm{Dirac}(\XX)}$. According to \thmref{thm:svc consistency}, in the infinite sample limit we would have
\begin{align}
    f(x)&=\int_\XX\lambda(\omega)t(\omega)\delta(x;\omega)\,dV(\omega),\\
    &=\lambda(x)t(x).
\end{align}
Since the norm of $f$ should be minimal and $f$ should satisfy $t(x)f(x) \geq 1$, we can deduce that $\lambda(x)=1$ for all $x$. We emphasize that when $x$ is on the decision boundary, it would not be participating in the optimization since we have no constraint in the program for those points. Consequently, if $t\in\ann{\HH(\Omega)}{\sigma(\XX)}$, then $t$ would remain feasible and would dominate all other $f\in\ann{\HH(\Omega)}{\sigma(\XX)}$ in terms of the training loss and the regularization score. 
\end{proof}

\subsection{\Thmref{thm:ann transfer}}
\begin{proof}
Using \lemref{lem:subset}, we know that both classes are equal to $\HH(\XX)$. Since $\HH(\XX)$ is normal, we know that with probability $1-\delta$ it is true that
\begin{equation}
    |f(x)-\hat{f}(x)|<\epsilon.
\end{equation}
Moreover, since $\partial\HH(\XX)$ is normal as well, it is true that
\begin{equation}
    \|\nabla f(x)-\nabla\hat{f}(x)\|<\epsilon'.
\end{equation}
Given both conditions, we can infer that a gradient based attack would follow a similar path for both $f$ and $\hat{f}$. 
\end{proof}

\subsection{\Thmref{thm:not normal}}
\begin{proof}
It is enough to find a learning rule $\AAA$ with respect to $L^2(\XX)$ that does not satisfy the condition for being normal. The SVC learning rule for $\ann{L^2(\XX)}{\mathrm{Dirac}(\XX)}$ is such a learning rule. To see why, imagine a set of training points $S\subset\XX$. Then $f$ would be a point mass function. Since the point masses are not measurable inside $\XX$, then it is almost surely true that
\begin{equation}
    |f(x)-t(x)| \geq 1.
\end{equation}
Thus, $L^2(\XX)$ is not a normal hypothesis class.
\end{proof}

\subsection{\Thmref{thm:transfer}}
\begin{proof}
First, assume that $\AAA$ is a nonuniform learner with respect to $A^2(\XX)$. By the definition of nonuniform learnability, for any $\epsilon,\delta>0$ a natural number $N$ exists that for any countable training set $S$ larger than $N$ and with probability $1-\delta$ we have that
\begin{equation}
    \ell(t(x),x,\AAA(S))\leq\epsilon.
\end{equation}

Consider all sequences $\seq{S_n}{n=1}{\infty}$ of dense training sets $S_n$ in which
\begin{equation}
    \lim_{n\rightarrow\infty}\,S_n=\XX.
\end{equation}
Then every $\seq{\AAA(S_n)}{n=1}{\infty}$ is converging compactly to $\AAA(\XX)$. We can see that any countable training set would appear in at least one of the sequences that we are considering. Consequently, for any countable training set $S$ larger than $N$ and with probability $1-\delta$ we have that
\begin{equation}
\label{eqn:val transfer}
    |\AAA(S)(x)-\AAA(\XX)(x)|\leq\epsilon.
\end{equation}

\begin{proposition}
\label{prop:ccd}
Suppose $\{f_n\}_{n=1}^\infty\subset\mathcal{O}(\XX)$ converges uniformly on compact subsets of $\XX$ to the function $f:\XX\rightarrow\mathbb{C}$. Then $f\in\mathcal{O}(\XX)$ and for each $\alpha\in\mathbb{N}^n$,
\begin{equation}
\lim_{n\rightarrow\infty} \partial^\alpha f_n=\partial^\alpha f    
\end{equation}
compactly in $\XX$. $\partial^\alpha$ is a shorthand notation for
\begin{equation}
    \frac{\partial^{|\alpha|}}{\partial x_1^{\alpha_1}\partial x_2^{\alpha_2}\cdots\partial x_d^{\alpha_d}}, \quad |\alpha|=\sum_{j=1}^d\alpha_j.
\end{equation}
\end{proposition}
Thus, we can deduce that
\begin{equation}
\label{eqn:del transfer}
    |\frac{\partial\AAA(S)}{\partial x_j}(x)-\frac{\partial\AAA(\XX)}{\partial x_j}(x)|\leq\epsilon\quad j=1,\cdots,d.
\end{equation}
\Eqnref{eqn:val transfer} and \eqnref{eqn:del transfer} show that $A^2(\XX)$ and $\partial A^2(\XX)$ are normal hypothesis classes with respect to nonuniform learners.

\begin{proposition}[\cite{Krantz2010OnLO}]
\label{prop:consistent transfer}
Let $\seq{f_n}{n=1}{\infty}$ be a sequence of holomorphic functions on a domain $\XX\subset\mathbb{C}^d$. Assume that the sequence converges pointwise to a limit function $f$ on $\XX$. Then $f$ is holomorphic on a dense open subset of $\XX$. Also the convergence is uniform on compact subsets of the dense open set.
\end{proposition}
\Propref{prop:consistent transfer} shows that when $\AAA$ is a consistent learner for $A^2(\XX)$ and $\seq{\AAA(S_n)}{n=1}{\infty}$ is converging pointwise to $\AAA(\XX)$, then for some dense open $K\subset\XX$ and with probability $1-\delta$ we have that
\begin{align}
    |\AAA(S)(x)-\AAA(\XX)(x)|\leq\epsilon&\quad x\in K\\
    |\frac{\partial\AAA(S)}{\partial x_j}(x)-\frac{\partial\AAA(\XX)}{\partial x_j}(x)|\leq\epsilon&\quad j=1,\cdots,d.
\end{align}
Consequently, we can find a similar bound for all $x\in\XX$ with some error probability $\delta' \propto \delta$. As a result, $A^2(\XX)$ and $\partial A^2(\XX)$ are normal hypothesis classes with respect to universally consistent learners.
\end{proof}

\subsection{\Thmref{thm:paradox}}
\begin{proof}
The theorem is a direct consequence of \propref{prop:consistent transfer}.
\end{proof}

\subsection{\Thmref{thm:dirichlet}}
\begin{proof}
Consider a smooth path $\gamma:[0,1]\rightarrow\XX$. The length of the image of this curve under $f\in A^2(\XX)$ is
\begin{equation}
    \|f(\gamma)\|=\int_0^1 \|\nabla f(\gamma(t))\|\gamma'(t)\,dt,
\end{equation}
Hence, when $\AAA$ has minimized $E[f]$, then it has minimized $\|f(\gamma)\|$ for all $\gamma$. Consequently, the image of all of the paths that start from a training sample $z$ would stay as close to $f(z)$ as possible. It follows that $f(\XX)$ and the imaginary axis, which represents the decision boundary, would have the minimum number of intersection points allowed by the training set $S$.

According to definition \ref{dfn:robust learner}, to show that $\AAA$ is robust, we have to show that for any $\epsilon \geq 0$ a non-negative integer $N$ exists for which the output of $\AAA$ for any two training set $S$ and $S'$ larger than $N$ would satisfy
\begin{equation}
\label{eqn:robust_condition}
    \sup_\XX |\AAA(S)-\AAA(S')|\leq\epsilon.
\end{equation}

Without loss of generality, assume that the labeling function $t$ partition $\XX$ into $M$ regions. Thus, as long as $N$ is big enough so that we have at least one sample from each region, the number of times that $f_n(\XX)$ intersects the decision boundary would not change. Consequently, adding any more samples to the training set would only results into a more accurate estimation of the position of the decision boundary. In other words, it is always the case that after adding the required $M$ support vectors to $S_n$, the position of the decision boundary would not make any drastic changes. As a result, for big enough training sets, \eqnref{eqn:robust_condition} would be satisfied.

Keep in mind that the same could not be said about a learning rule that does not minimize $E[f]$ since there is no guarantee that $f_n(\XX)$ and the decision boundary does not intersect more than necessary.
\end{proof}

\subsection{\Thmref{thm:ell2}}
\begin{proof}
$f$ has a representation
\begin{equation}
    f(z)=\sum_{\alpha\geq0}a_\alpha\varphi_\alpha(z)=a^H\varphi(z).
\end{equation}
We can prove the theorem by simply calculating $E[f]$, 
\begin{align}
    \int_\Omega\|\nabla f(z)\|^2\,dV(z)&=\int_\Omega a^HJ(z)J(z)^Ha\,dV(z),\\
    &=a^H\int_\Omega J(z)J(z)^H\,dV(z)a,\\
    &=a^H\Sigma a= a^Ha,
\end{align}
in which $J(z)$ is the Jacobian of the feature vector $\varphi(z)$.
\end{proof}

\subsection{\Thmref{thm:pde}}
\begin{proof}
The Lagrangian for the real-valued robust SVC learning rule is
\begin{align}
    \mathcal{L}&=\frac{1}{2}\int_\Omega \|\nabla h(\omega)\|^2\,dV(\omega)+\sum_{n=1}^N\lambda_n\big(1-t_n\int_\Omega h(\omega)s(x_n;\omega)\,dV(\omega)\big),\\
    &=\int_\Omega \frac{1}{2}\sum_{j=1}^dh_j(\omega)^2-\sum_{n=1}^N\lambda_nt_nh(\omega)s(x_n;\omega)\,dV(\omega)+\sum_{n=1}^N\lambda_n,\\
    &=\int_\Omega L(\omega,h,\nabla h)\,dV(\omega)+\sum_{n=1}^N\lambda_n.
\end{align}
By the Euler-Lagrange equations for a function of several variables, $h$ satisfies the following equation
\begin{equation}
    \frac{\delta L}{\delta h}-\sum_{j=1}^d\frac{\partial}{\partial \omega_j}\frac{\delta  L}{\delta h_j}=0.
\end{equation}
Consequently, $h$ must satisfy the following PDE
\begin{equation}
    -\Delta h(\omega)=\sum_{n=1}^N\lambda_nt_ns(x_n;\omega).
\end{equation}
\end{proof}

\subsection{\Thmref{thm:final}}
\begin{proof}
We know that
\begin{align}
    h(\omega)&=\sum_{n=1}^Na
    _n\mathfrak{s}_n(\omega),\\
    -\Delta \mathfrak{s}(x;\omega)&=\mathfrak{s}(x;\omega).
\end{align}
Computing $E[h]$, we would have
\begin{align}
    E[h]&=\int_\Omega \nabla h(\omega)\cdot\nabla h(\omega)\,dV(\omega),\\
    &=\sum_{n=1}^N\sum_{m=1}^Na_na_m\int_\Omega \nabla \mathfrak{s}_n(\omega)\cdot\nabla \mathfrak{s}_m(\omega)\,dV(\omega).
\end{align}

Since $\mathfrak{s}$ is twice differentiable, we may apply the divergence theorem to get
\begin{align}
\label{eqn:final}
    \int_\Omega \nabla\cdot\big(\mathfrak{s}_n(\omega)\nabla\mathfrak{s}_m(\omega)\big)\,dV(\omega)&=\int_\Omega \nabla\mathfrak{s}_n(\omega)\cdot\nabla\mathfrak{s}_m(\omega) - \mathfrak{s}_n(\omega)\mathfrak{s}_m(\omega)\,dV(\omega),\\
    &=\int_{\partial\Omega} \mathfrak{s}_n(\omega)\frac{\partial \mathfrak{s}_m(\omega)}{\partial n}\, dt,
\end{align}
When $\mathfrak{s}_m(\omega)$ satisfies either the natural Newman or Dirichlet boundary conditions, the right hand side of \eqnref{eqn:final} must vanish. Consequently,
\begin{equation}
    \int_\Omega \nabla\mathfrak{s}_n(\omega)\cdot\nabla\mathfrak{s}_m(\omega)\,dV(\omega)=\int_\Omega \mathfrak{s}_n(\omega)\mathfrak{s}_m(\omega)\,dV(\omega).
\end{equation}
By replacing the expression in $E[h]$ we would have
\begin{equation}
    E[h]=\|h\|^2_{L^2(\XX)}=\sum_{n=1}^N\sum_{m=1}^Na_na_m\int_\Omega \mathfrak{s}_n(\omega)\mathfrak{s}_m(\omega)\,dV(\omega).
\end{equation}
\end{proof}

\section{Domains of holomorphy}
\label{sec:domains}
In this section, we provide a brief summary of domains of holomorphy and describe how such domains could be constructed for common applications of ANNs. In complex analysis, the main subject of study are holomorphic functions of a scalar $z\in\mathbb{C}$. Domains of holomorphic functions in this setting are simple objects that are described by the Riemann mapping theorem and its generalizations, which states that if $\XX$ is a simply connected domain in $\mathbb{C}$ and is not $\mathbb{C}$ itself, then a biholomorphic map between $\XX$ and the unit disk $\mathbb{D}$ exists. One way to show that some domain is a domain of holomorphy is by finding a biholomorphic map between that domain and another domain holomorphy. A biholomorphic map is a map that is holomorphic and has a holomorphic inverse. As we have shown in \secref{sc:robust}, unit disk is a domain of holomorphy. Thus, all simply connected domains in $\mathbb{C}$ are domains of holomorphy.

However, an analogue for the Riemann mapping theorem does not exist for several complex variables. Domains of holomorphy has a geometrical property known as pseudoconvexity. The formal description of pseudoconvexity is very technical and we believe that the formal definition is not useful to the audience. Instead, we will present the reader with some domains of holomorphy, and then describe a method for constructing new domains of holomorphy from those building blocks.

We already know that the unit disk $\mathbb{D}\subset\mathbb{C}$ is a domain of holomorphy, and how we can use biholomorphic mappings to find new ones. In higher dimensions, we can construct a domain of holomorphy using a Cartesian product of other domains of holomorphy. As a special case, the Cartesian product of $d$ disks is called a polydisk $D^d(c,r)$ centered on $c$ and with radius $r$ and is defined as
\begin{equation}
    D^d(c,r)=\{z\in\mathbb{C}^d\,|\,|z_j-c_j|\leq r_j, j=1,\cdots,d\}.
\end{equation}
The boundary of the polydisk $D^d(0,1)$ could be used to encode $[0,1]^d$. To do so, map each real dimension to a complex variable with 
\begin{equation}
    z_j=e^{i\pi x_j}.
\end{equation}
The complex exponential is a periodic function that maps the real line to the unit circle. This procedure could be used to complexify datasets like MNIST.

It is possible to find domains of holomorphy that cannot be constructed using a Cartesian product of lower dimensional domains. One such domain is the ball $B(c,r)$ centered on $c$ with radius $r$
\begin{equation}
    B(c,r)=\{z\in\mathbb{C}^d\,|\,\|z-c\|\leq r\}.
\end{equation}
The boundary of the ball could be used to encode correlated dimensions that has a constant magnitude such as one-hot encoded categorical data. The real data could be mapped to a complex variable as the case for polydisks, with the extra step of normalizing $z\in\mathbb{C}^d$ to have $r$ as the magnitude. It goes without saying that a Cartesian product of balls and polydisks is also a domain of holomorphy.

\section{Are complex numbers necessary?}
One might imagine that we could have described the same framework without ever mentioning the complex numbers. In this section, we discuss how complex analysis helps us in our analysis, and why we think that complex analysis has much more to offer.

First, we know from various papers in the literature of adversarial examples phenomenon that the issue would occur in most applications of machine learning. Thus, if we want to describe the phenomenon, we have to analyse it in a context that is free from the choice of the hypothesis class. In our opinion, an analysis based on learning theory has the best chance of fulfilling this requirement. 

However, if we want to apply learning theory, we need to describe the phenomenon in learning theory terms as well. Consequently, we need to come up with definitions that conform to how learning theory defines its objects of study; the language of PAC learnability and uniform convergence.

The first obstacle the we would face is that real differentiable functions are not closed under uniform convergence. In other words, it is possible to find a sequence of real differentiable functions $\seq{f_n}{n=1}{\infty}$ that is uniformly converging to $f$, and yet $f$ is nowhere differentiable. Consequently, when we are dealing with the output of a learning rule $\AAA$, we cannot assume that the output is not ill-behaved in its derivatives. We recommend that the reader take a look at the Weierstrass function to get a picture of how ill-behaved the derivatives could become. This is a big issue for our analysis given that most of the definitions in the literature around the context of adversarial examples requires differentiation in one way or the other.

To guarantee that the pointwise limit of a sequence of differentiable functions is differentiable, the sequence needs to be pointwise converging in its value and uniform converging in its derivatives. According to PAC learnability, to achieve uniform convergence in derivatives, we need to train the derivatives of the hypothesis. But, how would we generate a training set for the gradient of the label "dog" with respect to the pixels in an image? We cannot ask a human to generate the derivatives! On its face, this is likely an impossible feat, and it seems that adversarial examples phenomenon is out of the reach of learning theory.

 This is where complex numbers show their true potential. It is known that sequences of complex differentiable functions (holomorphic functions) are closed under uniform convergence; \propref{prop:ccd} is a testament to this fact. In other words, if $\seq{f_n}{n=1}{\infty}$ is a sequence of holomorphic functions and the sequence is compactly converging to a function $f$, we know that $f$ is holomorphic. As we have stated in the main article, only complex-valued functions of a complex variable can be holomorphic. As a result, moving away from the complex number system to the real number system would be a huge step for some of the proofs in this paper. Nevertheless, while it would be more involved, it is probably possible to find similar theorems for real-valued hypotheses with the help of distributional derivatives.
 
 Apart from the ease of analysis of sequences of holomorphic functions, these functions are also unique in their geometrical properties. In the main article, we freely talk about the angles and lengths in the domain and the range space of a complex-valued hypothesis. We would not be able to do so if it was not for the holomorphicity of the hypothesis. The boundary tilting perspective of \cite{tanay2016boundary} is a good example of the steps needed to be taken for setting up a framework for rigorous study of the phenomenon in a geometrical sense. As demonstrated by \cite{tanay2016boundary}, coming up with alternative formal definitions for the real and the imaginary axis in our proposal is not a walk in the park. Moreover, even when \cite{tanay2016boundary} managed to do so, it proved to be too difficult to apply the geometrical intuition to a nonlinear hypothesis. In contrast, the geometrical interpretability of holomorphic functions enable us to circumvent these problems, and to translate our geometrical intuition to formal mathematical expressions with ease.
 
 In conclusion we argue that while it is very possible to recreate our framework without ever mentioning the complex number system, doing so would need even more exotic mathematical objects, and a much more involved discussion.
\end{document}